\documentclass[pmlr]{jmlr}

\RequirePackage{graphicx}
\usepackage{booktabs}
\usepackage{longtable}
\usepackage{amsmath}
\usepackage{multirow}
\usepackage{subcaption}
\usepackage{caption}
\usepackage{natbib}
\newcommand{\xhdr}[1]{{\vspace{1pt}\noindent\bfseries #1}.}
\usepackage{wrapfig}
\usepackage[table]{xcolor}
\graphicspath{{figure/}}

\makeatletter
\def\set@curr@file#1{\def\@curr@file{#1}}
\makeatother
\usepackage[load-configurations=version-1]{siunitx}

\jmlrproceedings{PMLR}{Proceedings of Machine Learning Research} \jmlrvolume{340} \jmlryear{2026} \jmlrworkshop{Machine Learning for Healthcare}

\title[Rethinking LLM Covariate Integration]{LLM-Extracted Covariates for Clinical Causal Inference: Rethinking Integration Strategies}

\author{%
  \Name[L. Liu]{Lei Liu$^{1,3,4}$} \Email{lliu\_04@arcadia.edu}\\
  \Name[J. Chen]{Jialin Chen$^{2,*}$} \Email{jialin.chen@yale.edu}\\
  \Name[K. Macropol]{Kathy Macropol$^{1,*}$} \Email{macropolk@arcadia.edu}\\
  \addr $^{1}$Department of Computer Science and Mathematics, Arcadia University, Philadelphia, PA, USA\\
  \addr $^{2}$Department of Computer Science, Yale University, New Haven, CT, USA\\
  \addr $^{3}$Department of Biostatistics, Yale University, New Haven, CT, USA\\
  \addr $^{4}$Jiangsu University, Zhenjiang, CN
}

\begin{document}

\maketitle

\let\thefootnote\relax\footnotetext{$^{*}$Corresponding authors.}

\begin{abstract}
Causal inference from electronic health records (EHR) is fundamentally
limited by unmeasured confounding: critical clinical states such as
frailty, goals of care, and mental status are documented in
free-text notes but absent from structured data. Large language
models can extract these latent confounders as interpretable,
structured covariates, yet how to effectively integrate them into
causal estimation pipelines has not been systematically studied.
Using the MIMIC-IV database with 21,859 sepsis patients, we compare
seven covariate-integration strategies for estimating the effect of
early vasopressor initiation on 28-day mortality, spanning
tabular-only baselines, traditional NLP representations, and three
LLM-augmented approaches. A central finding is that not all
integration strategies are equally effective: directly augmenting
the propensity score model with LLM covariates achieves the best
performance, while dual-caliper matching on text-derived categorical
distances restricts the donor pool and degrades estimation. In
semi-synthetic experiments with known ground-truth effects,
LLM-augmented propensity scores reduce estimation bias from 0.0143
to 0.0003 relative to tabular-only methods, and this advantage
persists under substantial simulated extraction error. On real
data, incorporating LLM-extracted covariates reduces the estimated
treatment effect from 0.055 to 0.027, directionally consistent
with the CLOVERS randomized trial, and a doubly robust estimator
yielding 0.031 confirms the robustness of this finding. Our
results offer practical guidance on when and how text-derived
covariates improve causal estimation in critical care. Our code is available at \url{https://github.com/fpxlei/LLM-Covariates-Causal}.
\end{abstract}

\section{Introduction}
\label{sec:intro}

Sepsis is a leading cause of mortality in intensive care units, and
the optimal timing of vasopressor initiation remains a subject of
ongoing clinical debate~\citep{Evans2021, Singer2016, Permpikul2019}.
The 2021 Surviving Sepsis Campaign guidelines recommend vasopressors
to maintain a mean arterial pressure above 65\,mmHg within the first
hour of care~\citep{Evans2021}, yet the evidence base remains mixed.
Observational studies have reached conflicting conclusions: some
reported that earlier norepinephrine reduced
mortality~\citep{Bai2014, OspinaTascon2020, Xu2022}, while others
found delayed initiation was associated with worse
outcomes~\citep{Black2020, ColonHidalgo2020}, and Waechter et
al.~\citep{Waechter2014} identified complex fluid-vasopressor
interactions. Randomized trials have not resolved the debate: the
CENSER trial~\citep{Permpikul2019} showed early norepinephrine
reduced shock control time, but the larger CLOVERS
trial~\citep{Self2023} found no significant mortality difference.
This persistent uncertainty motivates the use of observational EHR
data to complement trial evidence, but the validity of such analyses
hinges on adequately controlling for confounders that influence both
treatment decisions and outcomes.

A fundamental limitation of EHR-based causal inference is that
structured tabular data---lab values, vital signs, diagnosis
codes---capture only a fraction of the clinical state driving
treatment
decisions~\citep{Schneeweiss2009, Hernan2016, Yadav2018}. Critical
factors such as baseline functional
status~\citep{Muscedere2017, Bagshaw2014}, delirium
severity~\citep{Eidelman1996}, goals-of-care
preferences~\citep{Burns2016}, and infection source
heterogeneity~\citep{Stortz2022, OBrien2007} are extensively
documented in free-text clinical notes but absent from structured
fields. This \emph{unmeasured confounding} biases treatment effect
estimates and limits the credibility of observational
findings~\citep{Hager2024, Austin2011, Rubin1974}. Even
sophisticated approaches such as high-dimensional propensity score
adjustment~\citep{Schneeweiss2009} and target trial
emulation~\citep{Hernan2016} cannot fully address confounders that
exist only in unstructured text. The result is a persistent gap
between the clinical reality captured in notes and the statistical
models built from structured data alone.

Recent advances in large language models (LLMs) have demonstrated strong
zero-shot capabilities for clinical information
extraction~\citep{Nori2023, Agrawal2022, Alsentzer2023,
Sivarajkumar2024}. Unlike traditional NLP approaches that produce
opaque embeddings~\citep{Alsentzer2019}, LLMs can output structured,
interpretable covariates directly from clinical narratives. Several
works have explored using text data to address confounding in
observational
studies~\citep{Veitch2020, Keith2020, Feder2022, Roberts2020}.
However, a critical gap remains: \emph{given that LLMs can extract
clinically meaningful covariates from notes, how should these
covariates be integrated into a causal inference pipeline to most
effectively reduce confounding bias?}

In this work, we address this question through a systematic empirical
evaluation. Using the MIMIC-IV database~\citep{Johnson2023}, we
study the effect of early vasopressor initiation on 28-day mortality
in a cohort of 21,859 sepsis patients meeting Sepsis-3
criteria~\citep{Singer2016}. We extract seven structured clinical
covariates from discharge summaries using an LLM as a zero-shot
feature extractor, and compare seven strategies for integrating
these covariates and alternative text representations into causal
estimation. We validate estimates using semi-synthetic experiments
with known treatment effects and robustness tests under simulated
extraction noise, assess sensitivity via E-value
analysis~\citep{VanderWeele2017}, and investigate heterogeneous
treatment effects using causal forests~\citep{Wager2018} with
LLM-derived covariates as candidate effect modifiers. Because clinical causal inference demands transparency,
auditability, and long-term reproducibility that proprietary APIs
cannot guarantee, we additionally fine-tune an open-source model
(Qwen3-14B)~\citep{Qwen3} on multi-model consensus labels---labels
obtained by majority vote across three frontier LLMs (GPT-4o,
Gemini-2.5-Pro, Claude Sonnet 4). The fine-tuned model achieves
higher extraction accuracy than the proprietary baseline.

\xhdr{Generalizable Insights} Our evaluation yields three insights relevant to the broader
ML-for-health community:

\begin{enumerate}
    \item \textbf{Integration strategy matters.} We provide empirical
    evidence on whether the method of incorporating text-derived
    covariates---direct propensity score augmentation, two-stage
    matching, or inverse probability weighting---meaningfully affects
    treatment effect estimates, or whether the primary value lies in
    simply having access to these covariates.

    \item \textbf{Interpretable extraction outperforms black-box
    embeddings.} By comparing LLM-extracted structured covariates
    against BioClinicalBERT embeddings~\citep{Alsentzer2019}, we
    show that interpretable covariates achieve lower bias in
    semi-synthetic experiments while remaining clinically auditable.
    We further fine-tune Qwen3-14B~\citep{Qwen3} as a locally
    deployable alternative that surpasses the proprietary extraction
    baseline, addressing data privacy concerns in clinical settings
    by enabling covariate extraction without transmitting protected
    health information to external APIs.

    \item \textbf{LLM-derived covariates enable novel heterogeneous
    treatment effect (HTE) discovery.}
    We demonstrate that text-derived clinical concepts such as
    functional status and goals of care can serve as effect modifiers,
    identifying patient subgroups with differential treatment
    responses that are invisible to structured-data-only approaches.
\end{enumerate}

\section{Related Work}
\label{sec:related}

\paragraph{Causal inference from EHR data.}
Propensity score methods are the standard approach for estimating
treatment effects from observational EHR
data~\citep{Austin2011, Rubin1974, Schneeweiss2009}, and have been
widely applied to study vasopressor timing in critical
care~\citep{Bai2014, OspinaTascon2020, Waechter2014, Xu2022,
ColonHidalgo2020, Permpikul2019, Self2023}. A widely acknowledged
limitation across these studies is unmeasured confounding: structured
EHR fields omit critical clinical states such as
frailty~\citep{Muscedere2017, Bagshaw2014}, goals of
care~\citep{Burns2016}, mental status~\citep{Eidelman1996}, and
infection source heterogeneity~\citep{Stortz2022, OBrien2007} that
drive treatment decisions. Hern\'an and Robins~\citep{Hernan2016}
emphasized target trial emulation for observational causal inference,
and Schneeweiss et al.~\citep{Schneeweiss2009} proposed
high-dimensional propensity scores, but neither approach can capture
confounders that exist only in unstructured text. More broadly,
causal estimation under partial or systematically missing covariates
has been studied in settings where key confounders are incompletely
observed~\citep{Parbhoo2018partial, Parbhoo2020bottleneck}, and
causal inference frameworks have been applied to sepsis
detection~\citep{Li2024cisepsis}. Recent benchmarks have begun
evaluating LLM-derived features in clinical prognostic
tasks~\citep{Wang2026react}, though the integration of such
features into causal estimation pipelines remains unexplored.
Our work directly targets this gap.

\paragraph{NLP for confounding adjustment.}
Several lines of work have explored using text-derived features to
reduce confounding in observational studies. Veitch et
al.~\citep{Veitch2020} proposed adapting text embeddings for causal
inference, showing that document representations can serve as proxies
for unmeasured confounders in a semi-supervised framework. Keith et
al.~\citep{Keith2020} provided a comprehensive review of methods for
using text to remove confounding, identifying key challenges
including the choice of text representation and the assumption that
text captures all relevant confounders. Feder et
al.~\citep{Feder2022} presented a broader survey situating causal
inference within the NLP landscape, covering settings where text
serves as outcome, treatment, or confounder. Roberts et
al.~\citep{Roberts2020} demonstrated text matching for confounding
adjustment in political science, establishing that controlling for
text can substantially reduce bias when relevant confounders are
expressed in documents. Weld et al.~\citep{Weld2022} provided an
empirical evaluation framework for text-based confounding adjustment,
highlighting challenges in distinguishing genuine bias reduction from
overfitting. Pryzant et al.~\citep{Pryzant2021} studied causal
effects of linguistic properties, introducing methods for estimating
effects when treatments are embedded in
text. Yadav et al.~\citep{Yadav2018} surveyed EHR mining approaches
including text-based feature extraction for clinical prediction and
decision-making. Most existing approaches rely on dense embeddings
from pretrained language models such as
BioClinicalBERT~\citep{Alsentzer2019}, which capture semantic
information but produce opaque feature vectors that are difficult to
validate
clinically~\citep{Keith2020, Veitch2020, Feder2022}. Closer to our
work, several studies have focused on interpretable text-derived
confounders for clinical causal inference---uncovering confounders
from oncology notes~\citep{zeng2022uncovering}, integrating EHR
text across imputation and matching~\citep{mozer2023leveraging},
constructing substitute confounders via probabilistic factor
models~\citep{zhang2019medical}, and using LLM classification with
measurement error correction to recover unobserved
confounders~\citep{lee2024controlling}. Our work differs by
comparing black-box embeddings against LLM-extracted
\emph{interpretable} covariates, and by systematically evaluating
multiple integration strategies rather than proposing a single
method.

\paragraph{LLMs for clinical information extraction.}
Large language models have demonstrated strong performance on medical
reasoning benchmarks, with GPT-4 achieving approximately 91\% on
USMLE-style questions~\citep{Nori2023}. Hager et
al.~\citep{Hager2024} further evaluated LLM limitations in clinical
decision-making, identifying persistent challenges in reliability
and calibration. In clinical NLP, LLMs have been applied to a broad
range of extraction tasks. Agrawal et al.~\citep{Agrawal2022} showed
that LLMs are effective few-shot clinical information extractors,
matching or exceeding supervised baselines with minimal labeled data.
Yang et al.~\citep{Yang2022} developed GatorTron, a large clinical
language model trained on over 82 billion words of clinical text,
demonstrating strong performance on clinical NLP benchmarks.
Alsentzer et al.~\citep{Alsentzer2023} demonstrated zero-shot
phenotyping of postpartum hemorrhage from discharge notes using a
publicly available LLM, achieving high fidelity with interpretable
concept-level extraction. Sivarajkumar et
al.~\citep{Sivarajkumar2024} conducted a comprehensive evaluation
of prompting strategies for zero-shot clinical NLP across multiple
tasks and models, comparing GPT-3.5, LLaMA-2, and Gemini. Earlier
work on clinical text representations by Alsentzer et
al.~\citep{Alsentzer2019} introduced BioClinicalBERT, pretrained on
MIMIC-III clinical notes, which remains a widely used baseline for
clinical embedding tasks. Despite these advances in extraction
capability~\citep{Nori2023, Agrawal2022, Alsentzer2023,
Yang2022, Sivarajkumar2024}, existing work focuses on evaluating
\emph{whether} LLMs can extract clinical information accurately. The
question of \emph{how} these extracted variables should be integrated
into downstream statistical analyses---particularly causal inference
pipelines where covariate selection and balance directly affect
validity~\citep{Austin2011, Keith2020, Veitch2020}---remains largely
unexplored. Our contribution addresses this gap: we treat LLM
extraction as a feature engineering step within a causal inference
pipeline and systematically compare integration strategies on both
semi-synthetic and real clinical data.

\section{Methods}
\label{sec:methods}

\begin{figure}[t]
\centering
\includegraphics[width=\columnwidth]{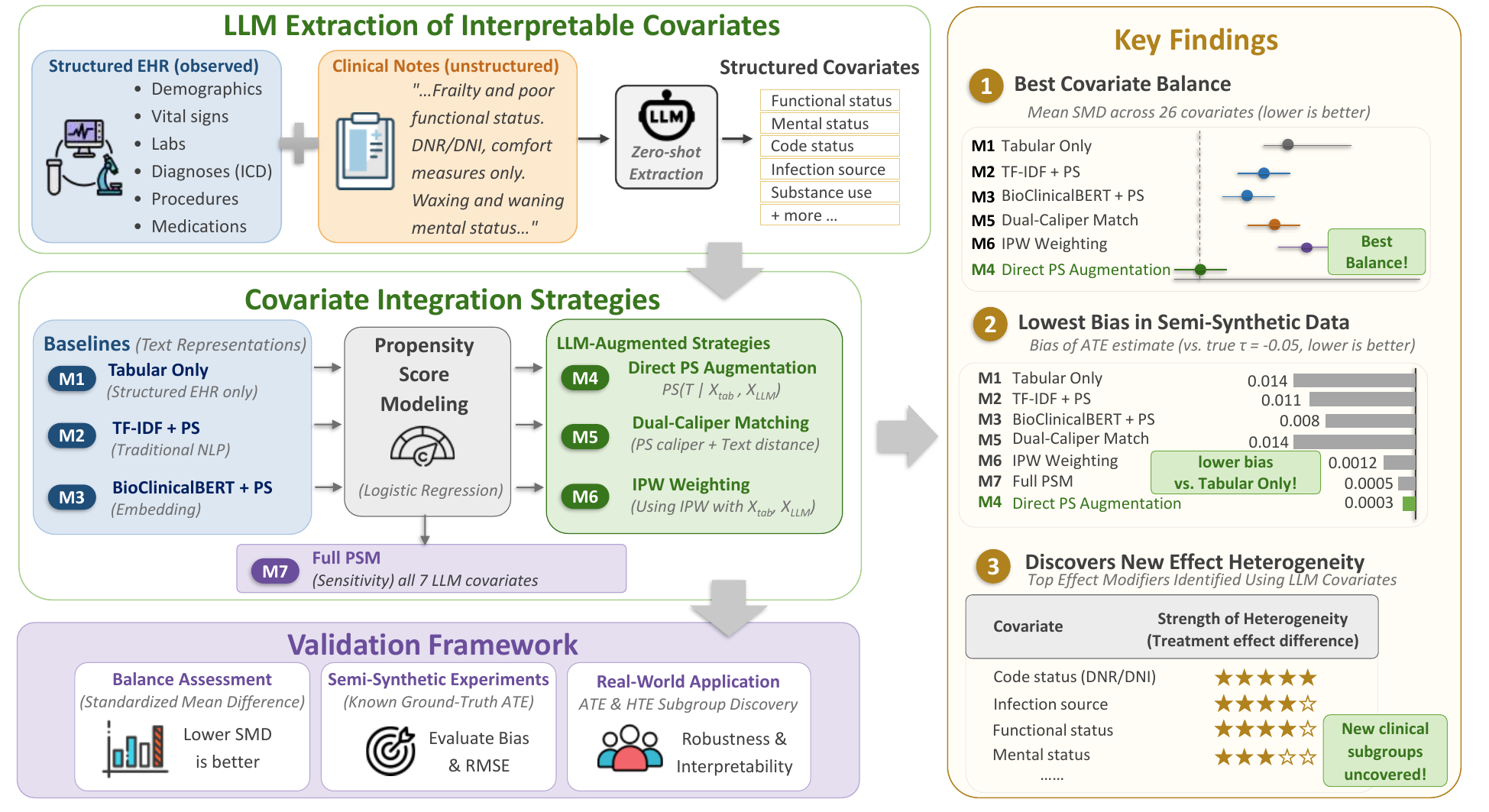}
\captionof{figure}{\textbf{Method overview and key findings.} \emph{Left:} We extract structured covariates from MIMIC-IV discharge summaries via zero-shot LLM extraction and compare seven integration strategies (M1--M3: baselines; M4--M6: LLM-augmented; M7: sensitivity analysis using all seven covariates). \emph{Right:} Direct propensity score augmentation (M4) achieves the best covariate balance (mean SMD 0.014 across 26 tabular covariates) and lowest bias in semi-synthetic experiments (0.0003 vs.\ true $\tau = -0.05$). LLM-extracted covariates also enable exploratory heterogeneous treatment effect discovery.}
\label{fig:pipeline}
\vspace{-1em}
\end{figure}

We introduce causal problem formulation in \S\ref{sec:formulation}. Our proposed approach consists of three stages: (1)~text-derived feature extraction via TF-IDF, BioClinicalBERT, or LLM-based structured extraction (\S\ref{sec:features}), (2)~seven causal estimation strategies spanning propensity score matching (PSM) and inverse probability weighting (IPW) (\S\ref{sec:estimation}), and (3)~validation through semi-synthetic benchmarks, E-value analysis, and heterogeneous treatment effect discovery (\S\ref{sec:validation}). Figure~\ref{fig:pipeline} demonstrates the overall workflow and summarizes our key empirical findings.

\subsection{Problem Formulation}
\label{sec:formulation}

We adopt the potential outcomes framework~\citep{Rubin1974}. Let
$T_i \in \{0,1\}$ denote whether patient $i$ received early
vasopressor initiation within 4\,hours of sepsis onset, $Y_i$
denote 28-day all-cause mortality, and $\mathbf{X}_i$ denote
observed baseline covariates. The average treatment effect (ATE) is
\begin{equation}
  \tau = \mathbb{E}\bigl[Y_i(1) - Y_i(0)\bigr]
\end{equation}
where $\mathbb{E}[\cdot]$ denotes expectation over the patient
population and $Y_i(t)$ is the potential outcome under treatment
$t$. Identification of $\tau$ requires consistency
($Y_i = Y_i(T_i)$), positivity
($0 < P(T_i = 1 \mid \mathbf{X}_i) < 1$ for all $\mathbf{X}_i$
in the support, where $P(\cdot)$ denotes probability), and
conditional ignorability
($\{Y_i(0), Y_i(1)\} \perp\!\!\!\perp T_i \mid
\mathbf{X}_i$)~\citep{Rubin1974, Rosenbaum1983}.

The key threat to conditional ignorability is that structured EHR
data omit clinically important confounders documented in free-text
clinical notes but absent from tabular fields. Let
$\mathbf{X}^{\text{tab}}_i$ denote the structured tabular
covariates and $\mathbf{X}^{\text{llm}}_i$ denote covariates
extracted from clinical text by an LLM. When only structured
covariates are available, the ignorability assumption is likely
violated:
\begin{equation}
  \{Y_i(0), Y_i(1)\} \not\!\perp\!\!\!\perp T_i \mid
  \mathbf{X}^{\text{tab}}_i
  \label{eq:violated}
\end{equation}
We hypothesize that augmenting structured covariates with
text-derived variables can better approximate conditional
ignorability:
\begin{equation}
  \{Y_i(0), Y_i(1)\} \perp\!\!\!\perp T_i \mid
  \bigl[\mathbf{X}^{\text{tab}}_i,\,
  \mathbf{X}^{\text{llm}}_i\bigr]
  \label{eq:augmented_ignorability}
\end{equation}
We define four feature sets: $\mathbf{X}^{\text{tab}}$ containing 26 structured EHR variables, $\mathbf{X}^{\text{tfidf}}$ containing TF-IDF representations, $\mathbf{X}^{\text{bert}}$ containing BioClinicalBERT embeddings, and $\mathbf{X}^{\text{llm}}$ containing LLM-extracted structured covariates. Among the LLM covariates, we distinguish $\mathbf{X}^{\text{llm}}_{\text{core}}$ (five confounders used in the primary analysis, M4--M6) from $\mathbf{X}^{\text{llm}}_{\text{all}}$ (all seven covariates, including two sensitivity variables used only in M7). See covariate details in Table~\ref{tab:llm_covariates}.

\subsection{Text-Derived Feature Extraction}
\label{sec:features}

We extract three classes of features from discharge summaries (processing details in Section~\ref{sec:data_extraction}).

\paragraph{TF-IDF representation.}
We apply L2-normalized term frequency--inverse document frequency weighting to discharge summary text, producing a sparse bag-of-words representation used as a baseline for text-based confounding adjustment.

\paragraph{BioClinicalBERT embeddings.}
We extract dense contextual embeddings from BioClinicalBERT~\citep{Alsentzer2019}, a transformer pretrained on MIMIC-III clinical notes, serving as a black-box embedding baseline.

\paragraph{LLM-extracted covariates.}
We prompt GPT-4o-mini as a zero-shot clinical feature extractor, which returns a structured JSON object containing seven pre-specified clinical covariates (as listed in Table~\ref{tab:llm_covariates}) We provide full prompts in Appendix~\ref{app:prompt}. Unlike the preceding representations, these covariates are fully interpretable and clinically auditable.

The final five core confounders selected for the primary analysis
are functional status~\citep{Muscedere2017, Bagshaw2014}, mental
status~\citep{Eidelman1996}, code status~\citep{Burns2016},
infection source~\citep{Stortz2022}, and substance use
history~\citep{OBrien2007}. Each was chosen because it plausibly
confounds the relationship between vasopressor initiation and
mortality, and is routinely documented in clinical notes but absent
from structured EHR fields. Two additional sensitivity
covariates---source control and family support---are included only
in M7 due to ambiguous temporal relationships to treatment.

\begin{table}[t]
\centering
\captionof{table}{LLM-extracted clinical covariates. Core
confounders enter the primary analysis M4--M6; all seven enter M7.
Code status abbreviations: DNR (do not resuscitate), DNI (do not
intubate), CMO (comfort measures only).}
\label{tab:llm_covariates}
\small
\begin{tabular}{llc}
\toprule
\textbf{Covariate} & \textbf{Categories} & \textbf{Role} \\
\midrule
Functional status  & independent, partial, fully dependent & Core \\
Mental status      & alert, confused, delirious, obtunded, comatose & Core \\
Code status        & full code, DNR, DNI, comfort measures & Core \\
Infection source   & pulmonary, abdominal, urinary, skin, blood, CNS & Core \\
Substance use      & none, alcohol, opioids, stimulants, multiple & Core \\
\midrule
Source control     & achieved, not achieved, N/A, pending & Sensitivity \\
Family support     & involved, limited, absent, conflicted & Sensitivity \\
\bottomrule
\end{tabular}
\vspace{-1em}
\end{table}

\subsection{Causal Estimation Strategies}
\label{sec:estimation}

We evaluate seven strategies in three groups
(Table~\ref{tab:methods_summary}). All propensity scores are
estimated via logistic regression~\citep{Rosenbaum1983} with
coefficients estimated by maximum likelihood. Matching uses 1:1
nearest-neighbor without replacement with a caliper of 0.2
standard deviations of the logit propensity
score~\citep{Austin2011}, with Abadie--Imbens
heteroskedasticity-robust standard errors~\citep{Abadie2016}.

\paragraph{Baselines (M1--M3).}
These methods do not use LLM-extracted covariates.
\textbf{M1} estimates propensity scores from
$\mathbf{X}^{\text{tab}}$ alone.
\textbf{M2} appends $\mathbf{X}^{\text{tfidf}}$ features.
\textbf{M3} appends $\mathbf{X}^{\text{bert}}$ embeddings.
The ATE is estimated as~\citep{Austin2011}:
\begin{equation}
  \hat{\tau}_{\text{PSM}} = \frac{1}{N_1}
  \sum_{i:\,T_i=1} \bigl[Y_i - Y_{j(i)}\bigr]
  \label{eq:psm}
\end{equation}
where $j(i)$ denotes the matched control for treated unit $i$
and $N_1$ is the number of treated patients.

\paragraph{LLM-augmented (M4--M6).}
\textbf{M4} directly appends
$\mathbf{X}^{\text{llm}}_{\text{core}}$ to the propensity score
model and estimates $\tau$ via Eq.~\eqref{eq:psm}.

\textbf{M5} uses a two-stage procedure: patients are first matched
on the tabular propensity score, then the final pair is selected to
minimize Hamming distance in the LLM-covariate space:
\begin{equation}
  j(i) = \mathop{\mathrm{arg\,min}}_{k \in \mathcal{C}(i)}\;
  d_{\text{H}}\!\bigl(\mathbf{X}^{\text{llm}}_{\text{core},i},\,
  \mathbf{X}^{\text{llm}}_{\text{core},k}\bigr)
  \label{eq:m5}
\end{equation}
where $d_{\text{H}}(\cdot,\cdot)$ denotes the Hamming distance
between two categorical vectors and $\mathcal{C}(i)$ is the set
of controls within the caliper of 0.2 standard deviations.

\textbf{M6} uses the same covariates as M4 but applies stabilized
inverse probability weighting~\citep{Robins2000}:
\begin{equation}
  \hat{\tau}_{\text{IPW}} = \frac{1}{N} \sum_{i=1}^{N}
  \left[
    \frac{T_i\, Y_i\, \hat{P}(T{=}1)}{\hat{e}_i}
    - \frac{(1 - T_i)\, Y_i\, \hat{P}(T{=}0)}
      {1 - \hat{e}_i}
  \right]
  \label{eq:ipw}
\end{equation}
where $N$ is the total number of patients, $\hat{e}_i$ is the
estimated propensity score for patient $i$, and $\hat{P}(T{=}1)$
denotes the marginal treatment probability used for stabilization.
Propensity scores are clipped to $[0.01, 0.99]$ and weights are
truncated at the 1st and 99th percentiles to mitigate extreme
values, followed by within-group renormalization. Variance is
estimated using robust sandwich standard errors. Further
implementation details are provided in
Appendix~\ref{app:implementation}.

\paragraph{Sensitivity (M7).}
\textbf{M7} repeats the M4 design with
$\mathbf{X}^{\text{llm}}_{\text{core}}$ replaced by
$\mathbf{X}^{\text{llm}}_{\text{all}}$, adding source control and
family support.

\begin{table}[ht]
\centering
\captionof{table}{Summary of the seven estimation strategies.}
\label{tab:methods_summary}
\small
\begin{tabular}{clll}
\toprule
& \textbf{Covariates} & \textbf{Estimator} & \textbf{Group} \\
\midrule
M1 & $\mathbf{X}^{\text{tab}}$ & PSM & Baseline \\
M2 & $\mathbf{X}^{\text{tab}} + \mathbf{X}^{\text{tfidf}}$ & PSM
  & Baseline \\
M3 & $\mathbf{X}^{\text{tab}} + \mathbf{X}^{\text{bert}}$ & PSM
  & Baseline \\
M4 & $\mathbf{X}^{\text{tab}} + \mathbf{X}^{\text{llm}}_{\text{core}}$
  & PSM & LLM-aug. \\
M5 & $\mathbf{X}^{\text{tab}}$ then
  $\mathbf{X}^{\text{llm}}_{\text{core}}$
  & Dual-caliper & LLM-aug. \\
M6 & $\mathbf{X}^{\text{tab}} + \mathbf{X}^{\text{llm}}_{\text{core}}$
  & IPW & LLM-aug. \\
M7 & $\mathbf{X}^{\text{tab}} + \mathbf{X}^{\text{llm}}_{\text{all}}$
  & PSM & Sensitivity \\
\bottomrule
\end{tabular}
\vspace{-1em}
\end{table}
\subsection{Validation Framework}
\label{sec:validation}

We assess credibility through three complementary evaluation strategies (Figure~\ref{fig:pipeline}).

\paragraph{Covariate balance assessment.}
We report standardized mean differences (SMD) across all covariates
before and after adjustment, with adequate balance defined as SMD
below 0.1 for each individual covariate~\citep{Austin2011}.

\paragraph{Semi-synthetic experiments.}
We construct semi-synthetic datasets that preserve the observed MIMIC-IV covariate structure while imposing known treatment assignment and outcome mechanisms with a pre-specified ground-truth ATE (Appendix~\ref{app:synthetic}). Each method's ability to recover the true effect is evaluated via bias, RMSE, and 95\% confidence interval coverage.

\paragraph{Real-world application.}
On the observed MIMIC-IV cohort, we estimate the ATE of early vasopressor initiation on 28-day mortality and assess robustness through E-value sensitivity analysis~\citep{VanderWeele2017}, which quantifies the minimum unmeasured confounder strength needed to explain away each observed effect. We further explore heterogeneous treatment effects using causal forests~\citep{Wager2018} with LLM-extracted covariates as candidate effect modifiers and tabular covariates as nuisance confounders, applying propensity score trimming to enforce overlap~\citep{Crump2009}. As a directional reference, we compare our estimates against the CLOVERS trial~\citep{Self2023}, a multicenter RCT of early vasopressor versus liberal fluid strategies in septic shock.

\section{Cohort}
\label{sec:cohort}

\subsection{Cohort Selection}
\label{sec:cohort_selection}
\begin{wrapfigure}{r}{0.5\columnwidth}
\centering
\includegraphics[width=0.5\columnwidth]{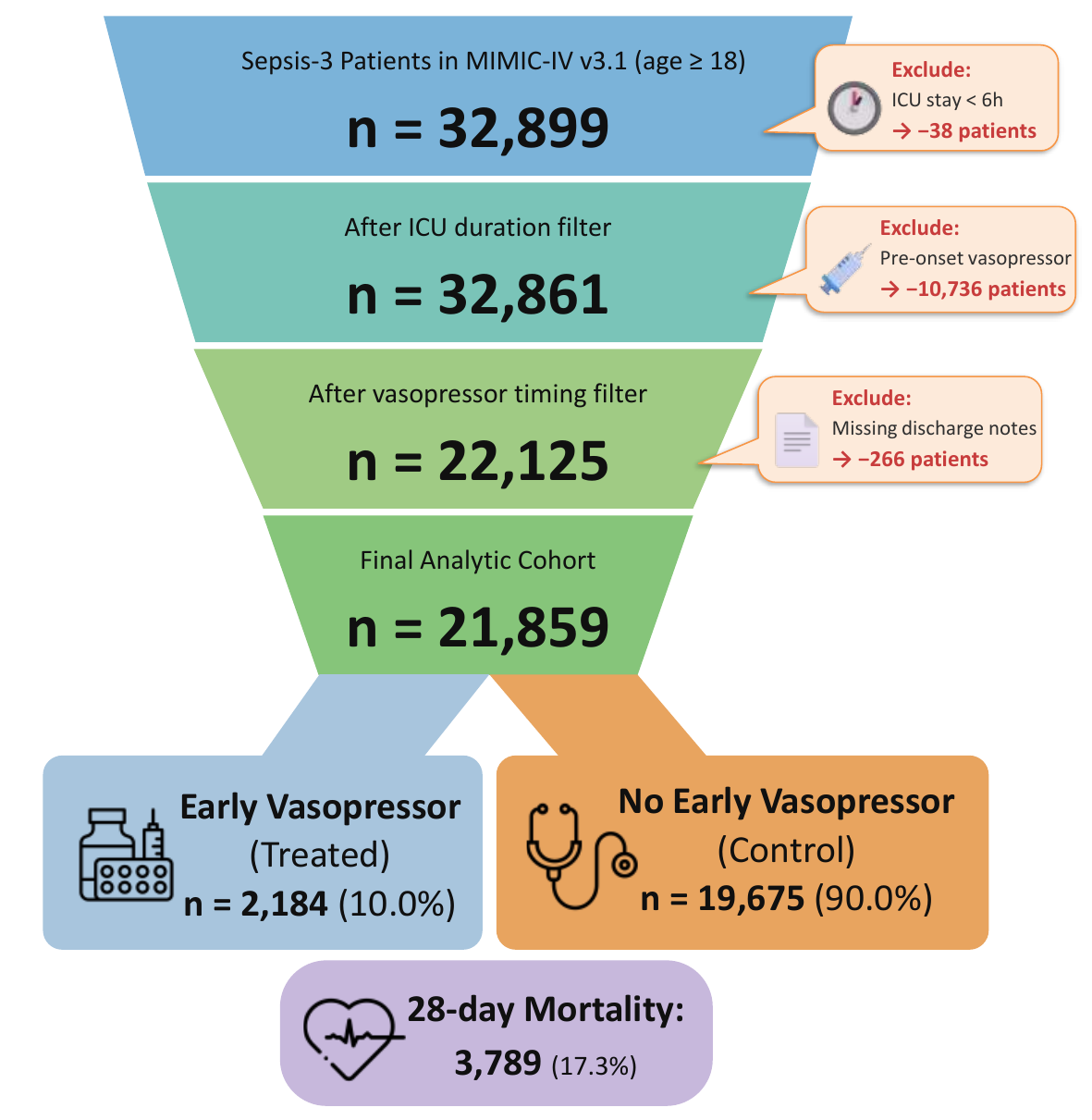}
\caption{Cohort selection flowchart. Starting from 32,899
Sepsis-3 patients, three exclusion criteria yield a final cohort of
21,859 patients.}
\label{fig:consort}
\end{wrapfigure}

We extract adult patients from
MIMIC-IV v3.1~\citep{Johnson2023} meeting Sepsis-3
criteria~\citep{Singer2016} and apply three sequential exclusion
criteria: ICU stay under 6 hours to avoid survivorship
bias~\citep{Seymour2017, Leisman2017}, vasopressor administration
prior to sepsis onset to restrict the study to incident
treatment~\citep{Waechter2014, Bai2014}, and missing discharge
summaries required for LLM-based extraction. As shown in
Figure~\ref{fig:consort}, the final cohort comprises 21,859
patients, of whom 2,184 received early vasopressors and 19,675
did not, with an overall 28-day mortality rate of 17.3\%.

\subsection{Data Extraction}
\label{sec:data_extraction}

\paragraph{Treatment and outcome.}
Early vasopressor initiation is defined as first administration of
norepinephrine, vasopressin, phenylephrine, epinephrine, or
dopamine within 4\,hours of sepsis onset, with 28-day all-cause
mortality as the outcome.

\paragraph{Structured covariates.}
We extract 26 baseline structured variables from MIMIC-IV derived
tables~\citep{Johnson2023}---demographics, severity scores,
first-day vital signs, laboratory values, and early
interventions---all measured at or before sepsis onset. Missing
values are imputed using multivariate imputation by chained
equations (MICE)~\citep{vanBuuren2011} with LLM-extracted
covariates as auxiliary variables.

\paragraph{Clinical text.}
Discharge summaries are extracted from MIMIC-IV-Note for all
21,859 patients; the extraction prompt restricts output to
pre-treatment and admission states (Appendix~\ref{app:prompt}).

As summarized in Table~\ref{tab:baseline}, treated patients had
substantially higher illness severity (Simplified Acute Physiology
Score II, SAPS-II: 47.9 vs.\ 37.6) and greater imbalance on
LLM-extracted covariates, consistent with confounding by
indication. Text-derived features are processed as follows: TF-IDF
is reduced to 50 dimensions via truncated SVD, BioClinicalBERT
embeddings are reduced from 768 to 50 dimensions via PCA, and
LLM-extracted covariates are one-hot encoded. Further
implementation details are in Appendix~\ref{app:implementation}.

\begin{table}[t]
\centering
\captionof{table}{Baseline characteristics by treatment group. Values are mean (SD) or $n$ (\%).}
\label{tab:baseline}
\small
\begin{tabular}{lccc}
\toprule
Characteristic & Treated ($n{=}2{,}184$) & Control ($n{=}19{,}675$) & SMD \\
\midrule
\multicolumn{4}{l}{\textit{Demographics \& Severity}} \\
~~Age, years & 66.9 (15.0) & 65.6 (16.9) & 0.081 \\
~~Female sex & 872 (39.9\%) & 8579 (43.6\%) & 0.075 \\
~~SOFA score & 3.6 (1.9) & 3.2 (1.6) & 0.280 \\
~~SAPS-II & 47.9 (15.9) & 37.6 (13.0) & 0.713 \\
~~Charlson index & 5.5 (2.8) & 5.4 (3.0) & 0.025 \\
\addlinespace
\multicolumn{4}{l}{\textit{Vital Signs}} \\
~~Heart rate, bpm & 88.9 (16.8) & 87.3 (16.3) & 0.099 \\
~~MAP, mmHg & 72.6 (7.9) & 78.1 (11.3) & 0.570 \\
~~Resp.\ rate, /min & 20.0 (4.1) & 19.9 (4.2) & 0.033 \\
~~Temp., $^{\circ}$C & 36.9 (0.7) & 36.9 (0.5) & 0.071 \\
~~SpO$_2$, \% & 97.0 (3.0) & 96.8 (2.1) & 0.080 \\
\addlinespace
\multicolumn{4}{l}{\textit{Laboratory Values}} \\
~~Creatinine, mg/dL & 2.1 (2.5) & 1.8 (1.9) & 0.123 \\
~~WBC, $10^3/\mu$L & 17.6 (15.1) & 14.0 (12.6) & 0.253 \\
~~Platelets, $10^3/\mu$L & 166.3 (111.3) & 190.8 (113.1) & 0.219 \\
~~Bilirubin, mg/dL & 2.7 (5.4) & 2.3 (5.3) & 0.067 \\
\addlinespace
\multicolumn{4}{l}{\textit{LLM-Extracted Confounders}} \\
~~Fully dependent & 277 (12.7\%) & 2,082 (10.6\%) & 0.066 \\
~~Altered mental status & 1,170 (53.6\%) & 10,186 (51.8\%) & 0.036 \\
~~Code status limitation & 430 (19.7\%) & 3,244 (16.5\%) & 0.083 \\
~~Pulmonary infection & 509 (23.3\%) & 5,591 (28.4\%) & 0.117 \\
~~Active substance use & 673 (30.8\%) & 6,192 (31.5\%) & 0.014 \\
~~28-day mortality & 573 (26.2\%) & 3,216 (16.3\%) & 0.243 \\
\addlinespace
\multicolumn{4}{l}{\textit{Outcome}} \\
~~28-day mortality & 573 (26.2\%) & 3216 (16.3\%) & 0.243 \\
\bottomrule
\end{tabular}
\vspace{-1em}
\end{table}

\subsection{Results on Synthetic Experiments}
\label{sec:synthetic}

We construct semi-synthetic datasets that preserve the real MIMIC-IV
covariates but impose known treatment and outcome mechanisms where
both depend on LLM covariates, creating unmeasured confounding by
construction for tabular-only methods
(Appendix~\ref{app:synthetic}). Over 200 replications with
$\tau = -0.05$, the LLM-augmented methods M4, M6, and M7
concentrate tightly around the true value
(Figure~\ref{fig:boxplot}), with M4 achieving the lowest bias of
0.0003 and 97.0\% coverage. The dual-caliper method M5 performs
poorly at 0.0137, comparable to the tabular-only baseline. M4
maintains lower bias than M1 even under 20\% simulated extraction
misclassification, confirming robustness to imperfect extraction
(Appendix~\ref{app:synthetic}, Table~\ref{tab:noise}). Full numerical
results are in Appendix~\ref{app:synthetic}.

\begin{center}
\begin{minipage}{\columnwidth}
\centering
\includegraphics[width=\columnwidth]{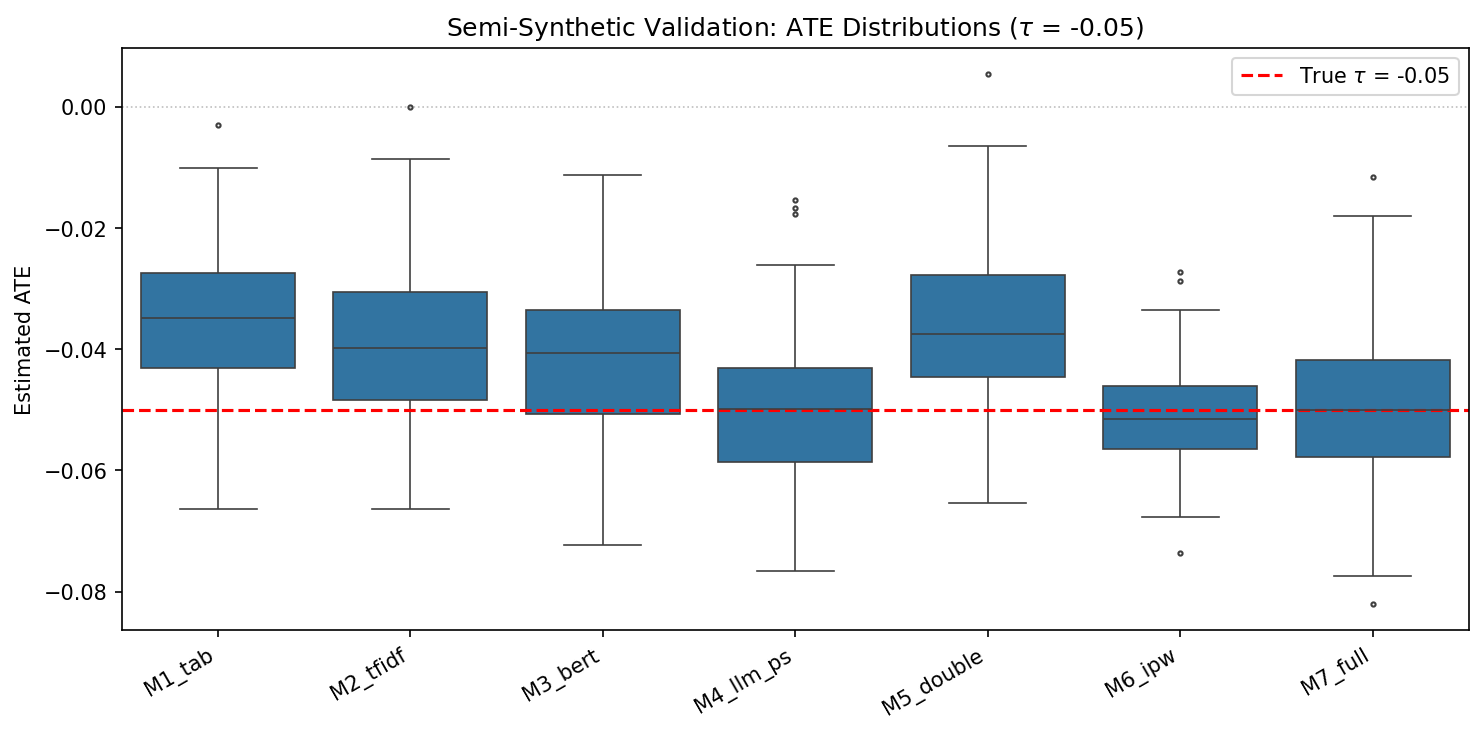}
\captionof{figure}{Distribution of ATE estimates across 200
simulations with true $\tau = -0.05$. LLM-augmented methods M4,
M6, and M7 concentrate tightly around the true value, while M1
shows substantial upward bias.}
\label{fig:boxplot}
\vspace{-1em}
\end{minipage}
\end{center}

\section{Results on Real Data}
\label{sec:results}


\subsection{LLM Extraction Quality}
\label{sec:extraction}

A reference standard for evaluating extraction accuracy is
established via majority vote across three frontier models---GPT-4o,
Gemini-2.5-Pro, and Claude Sonnet 4---applied to 3,200 discharge
summaries, with consensus exceeding 95\% for all covariates
(Appendix~\ref{app:figures}). As shown in
Table~\ref{tab:extraction}, the zero-shot GPT-4o-mini model used in
all primary analyses achieves a core mean accuracy of 55.6\%.
Fine-tuning Qwen3-14B on consensus labels substantially improves
extraction quality, raising core mean accuracy to 72.7\% and
yielding the largest gains on code status, from 65.7\% to 88.3\%,
and substance use, from 44.7\% to 85.5\%. As demonstrated by the
noise robustness analysis in Section~\ref{sec:synthetic}, even
imperfect extraction reduces confounding bias relative to omitting
these covariates entirely. The fine-tuned Qwen3-14B is released as
an open-source, locally deployable alternative to the proprietary
API.
\begin{table}[t]
\centering
\captionof{table}{Extraction accuracy against three-model consensus labels. GPT-4o-mini is evaluated zero-shot on all 3,200 samples; Qwen3-14B is evaluated on the held-out test set ($n = 640$) after LoRA fine-tuning. Core covariates (used in M4--M6) are above the line; sensitivity covariates (M7 only) are below.}
\label{tab:extraction}
\small
\setlength{\tabcolsep}{6pt}
\begin{tabular}{lcccc}
\toprule
& \multicolumn{2}{c}{\textbf{GPT-4o-mini}} & \multicolumn{2}{c}{\textbf{Qwen3-14B}} \\
\cmidrule(lr){2-3} \cmidrule(lr){4-5}
\textbf{Covariate} & \textbf{Acc.} & \textbf{F1} & \textbf{Acc.} & \textbf{F1} \\
\midrule
Code status         & 65.7\% & 0.495 & \textbf{88.3\%} & \textbf{0.709} \\
Infection source    & \textbf{65.9\%} & \textbf{0.573} & 63.1\% & 0.383 \\
Functional status   & 55.7\% & 0.542 & \textbf{72.5\%} & \textbf{0.547} \\
Mental status       & 46.2\% & 0.341 & \textbf{54.1\%} & 0.305 \\
Substance use       & 44.7\% & 0.291 & \textbf{85.5\%} & \textbf{0.447} \\
\midrule
\rowcolor{gray!20}
\textbf{Core mean}          & 55.6\% & 0.448 & \textbf{72.7\%} & \textbf{0.478} \\
\midrule
Family support      & 72.8\% & \textbf{0.375} & \textbf{80.9\%} & 0.335 \\
Source control      & 17.6\% & 0.206 & \textbf{57.3\%} & \textbf{0.285} \\
\midrule
\rowcolor{gray!20}
\textbf{Overall mean}        & 52.7\% & 0.403 & \textbf{71.7\%} & \textbf{0.430} \\
\bottomrule
\end{tabular}
\end{table}

\subsection{Covariate Balance}
\label{sec:balance}

\begin{figure}[t]
\centering
\includegraphics[width=\columnwidth]{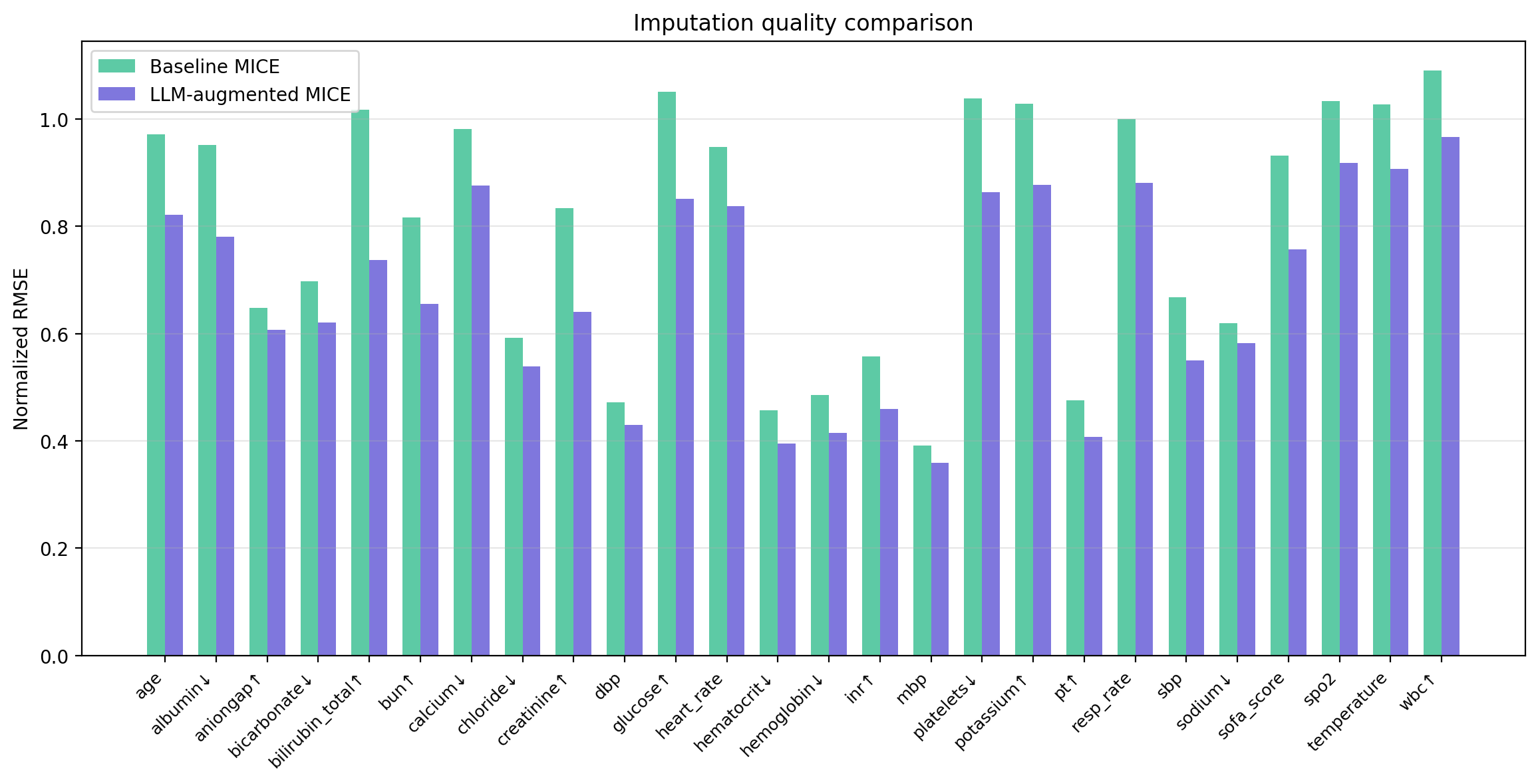}
\caption{Imputation quality across tabular covariates. Normalized
RMSE is computed via repeated holdout masking of observed entries:
for each covariate, 10\% of observed values are randomly masked
and imputed under each method, and RMSE against the true values is
normalized by each covariate's standard deviation. LLM-augmented
MICE reduces normalized RMSE on most variables relative to
standard MICE without auxiliary LLM covariates.}
\label{fig:rmse}
\vspace{-1em}
\end{figure}

Table~\ref{tab:balance} summarizes covariate balance across all
methods. M4 achieves the best balance among the primary methods on
tabular covariates, with mean SMD of 0.016 and all 26 covariates
balanced, as well as on LLM-extracted covariates with mean SMD of 0.013. The tabular-only
baseline leaves substantial imbalance on LLM covariates at mean SMD
of 0.058, confirming that these confounders are not controlled
without explicit text-derived adjustment. IPW achieves weaker
balance with only 18 of 26 tabular covariates meeting the threshold
and an effective sample size of 1,433 out of 2,184 treated.
Additionally, using LLM-extracted covariates as auxiliary variables
in MICE imputation improves imputation quality across most tabular
covariates (Figure~\ref{fig:rmse}). The full covariate-level love plot
(Figure~\ref{fig:love}) and LLM-covariate balance details are
provided in Appendix~\ref{app:figures}.

\begin{table}[t]
\centering
\captionof{table}{Covariate balance summary. Mean and maximum absolute SMDs over 26 tabular covariates. Balanced: SMD $< 0.1$.}
\label{tab:balance}
\small
\begin{tabular}{lccc}
\toprule
\textbf{Method} & \textbf{Mean SMD} & \textbf{Max SMD} & \textbf{Balanced} \\
\midrule
Unmatched         & 0.243 & 0.765 & 8/26 \\
M1: Tabular PSM   & 0.023 & 0.072 & 26/26 \\
M2: TF-IDF PSM    & 0.025 & 0.060 & 26/26 \\
M3: BERT PSM      & 0.023 & 0.065 & 26/26 \\
M4: LLM PSM       & 0.016 & 0.073 & 26/26 \\
M5: Dual-caliper  & 0.032 & 0.100 & 25/26 \\
M6: LLM IPW       & 0.093 & 0.383 & 18/26 \\
M7: LLM PSM (all) & \textbf{0.014} & \textbf{0.042} & 26/26 \\
\bottomrule
\end{tabular}
\vspace{-1em}
\end{table}


\subsection{Treatment Effect Estimates}
\label{sec:ate}

Table~\ref{tab:ate} presents the ATE estimates. All methods yield
a positive ATE, suggesting that early vasopressor initiation is
associated with increased 28-day mortality in this observational
cohort, though the magnitude varies substantially across methods.
The corresponding forest plot is provided in
Appendix~\ref{app:figures}.

The tabular-only baseline M1 estimates an ATE of 0.055,
indicating a 5.5 percentage point increase in mortality.
Incorporating LLM-extracted covariates into the propensity score
model via M4 reduces this to 0.036---a reduction of roughly one
third---suggesting that part of the apparent harm is
attributable to unmeasured confounding captured by the LLM
covariates. The TF-IDF method M2 produces the smallest and
non-significant estimate of 0.008 at $p = 0.53$, though this
may reflect noise from high-dimensional sparse features rather than
genuine confounding adjustment.

The positive direction of all estimates is broadly consistent with
the near-null finding of the CLOVERS trial~\citep{Self2023},
although the estimands differ in both outcome window and treatment
definition, and this comparison serves as a rough directional check
rather than formal validation.

As a robustness check, a doubly robust ATE estimated via
augmented inverse probability weighting (AIPW) with
ForestDRLearner and the same covariate set as M4 yields 0.031
with 95\% CI from $-$0.005 to 0.066, closely matching the M4
PSM estimate and confirming that our findings are not sensitive to
estimator choice. To verify that results are not driven by
post-treatment information leakage from discharge summaries, we
additionally restrict the input text to admission-time paragraphs
only---physically excluding hospital course and all post-discharge
sections---and obtain an M4 estimate of 0.026, directionally consistent with
the primary estimate of 0.036. E-values are reported in
Table~\ref{tab:ate}; computation details are in
Appendix~\ref{app:implementation}.

\begin{table}[ht]
\centering
\captionof{table}{Estimated ATE of early vasopressor initiation on 28-day mortality. Positive values indicate increased mortality.}
\label{tab:ate}
\small
\begin{tabular}{lcccc}
\toprule
\textbf{Method} & \textbf{ATE} & \textbf{95\% CI} & \textbf{$p$} & \textbf{E-value} \\
\midrule
M1: Tabular PSM & 0.055 & [0.030, 0.080] & ${<}0.001$ & 2.01 \\
M2: TF-IDF PSM & 0.008 & [$-$0.018, 0.034] & 0.530 & 1.28 \\
M3: BERT PSM & 0.038 & [0.012, 0.063] & 0.004 & 1.76 \\
M4: LLM PSM & 0.036 & [0.010, 0.061] & 0.006 & 1.73 \\
M5: Dual-caliper & 0.058 & [0.032, 0.083] & ${<}0.001$ & 2.04 \\
M6: LLM IPW & 0.051 & [0.029, 0.072] & ${<}0.001$ & 1.94 \\
M7: LLM PSM (all 7) & 0.027 & [0.001, 0.052] & 0.041 & 1.60 \\
AIPW (DR) & 0.031 & [$-$0.005, 0.066] & 0.087 & 1.66 \\
\bottomrule
\end{tabular}
\vspace{-1em}
\end{table}

\subsection{Heterogeneous Treatment Effects}
\label{sec:hte}

To explore treatment effect heterogeneity, we fit a CausalForestDML
model~\citep{Wager2018} with the five core LLM-extracted covariates
as candidate effect modifiers; implementation details are in
Appendix~\ref{app:implementation}.

Variable importance analysis identifies code status (0.28),
infection source (0.25), and functional status (0.21) as the three
most influential modifiers, collectively accounting for 74\% of
the total importance. Mental status (0.17) and source control
(0.10) contribute the remainder. All subgroup confidence intervals
include zero, and these results should be treated as exploratory.
Nonetheless, clinically plausible patterns emerge: patients with full code status show the smallest estimated effect
at 0.011 while those with DNR status show 0.035, and functionally
independent patients show a near-zero effect at 0.005 while fully
dependent patients show 0.032. These gradients illustrate
a key advantage of interpretable LLM-derived covariates: the
ability to define clinically meaningful subgroups invisible to
structured-data-only analyses. The variable importance plot and
detailed subgroup conditional average treatment effect (CATE)
tables are in Appendix~\ref{app:figures}.

\section{Discussion}
\label{sec:discussion}

This study systematically evaluates how LLM-extracted clinical covariates should be integrated into observational causal inference pipelines. Three key findings emerge.

\textbf{The simplest integration strategy is the most effective.} Among the three LLM-augmented approaches, directly incorporating structured covariates into the propensity score model achieves the best covariate balance and reduces the estimated ATE from 0.055 to 0.027, halving the effect magnitude. In contrast, the two-stage dual-caliper approach performs poorly, producing an ATE of 0.060 that exceeds even the tabular-only baseline. The underperformance likely reflects a fundamental trade-off: enforcing near-exact matching on categorical LLM covariates via Hamming distance severely restricts the donor pool, forcing acceptance of poorer matches on continuous tabular covariates. With only 2,184 treated patients and seven categorical variables, the combinatorial matching space is too constrained to simultaneously optimize both dimensions. IPW yields a reasonable point estimate but achieves weaker balance, with only 18 of 26 covariates meeting the 0.1 threshold. These results suggest that practitioners should prefer direct propensity score augmentation over more complex integration strategies.

\textbf{Interpretable LLM covariates outperform black-box embeddings.} In semi-synthetic experiments, the LLM-augmented propensity score method achieves bias of 0.0003 compared to 0.0082 for BioClinicalBERT embeddings---a reduction of over 95\%. In real data, the LLM method produces a smaller ATE that is more consistent with the near-null finding of the CLOVERS trial. The advantage of structured covariates likely stems from their direct alignment with clinical confounders: a variable explicitly encoding comfort-measures-only status captures a specific, well-defined confounder, whereas a 768-dimensional embedding distributes this signal across many latent dimensions, diluting its influence in the propensity score model.

\textbf{LLM-derived covariates reveal clinically plausible heterogeneity patterns.} Causal forest analysis identifies code status and functional status as the strongest effect modifiers---variables entirely absent from structured EHR data. While all subgroup effects remain statistically imprecise, the observed gradient is clinically plausible: functionally independent patients show a near-null effect, while fully dependent patients show a larger positive effect. These exploratory findings illustrate the unique value of text-derived covariates for generating hypotheses about treatment effect heterogeneity and warrant confirmatory investigation in larger, multi-center cohorts.

Taken together, our results suggest that the primary value of LLM extraction lies not in enabling sophisticated integration methods but in providing access to clinically meaningful confounders that are otherwise invisible. Even the simplest integration strategy yields substantial bias reduction when the right covariates are available. We emphasize that improved covariate balance is a necessary but not sufficient condition for valid causal inference; residual bias from unmeasured or post-treatment confounders may persist even when observed balance metrics appear favorable.

\paragraph{Limitations.}
Our study extracts covariates from discharge summaries, which are written after the clinical encounter and may capture post-treatment states despite prompt-level temporal restrictions; restricting extraction to admission or early nursing notes would more rigorously ensure pre-exposure measurement. Additionally, our static propensity score framework treats vasopressor initiation as a point exposure, whereas marginal structural models or target trial emulation~\citep{Hernan2016} would more appropriately handle the time-varying nature of this treatment decision. Finally, this is a single-center study using MIMIC-IV; external validation on multi-center datasets is needed to establish generalizability.

\acks{This work was supported by the National College Students' Innovation
and Entrepreneurship Training Program of Jiangsu University (Grant
No.~202510299093).}

\bibliography{references}

\newpage
\appendix

\section{LLM Extraction}
\label{app:prompt}

The following system prompt is used for all LLM-based covariate extraction. All calls use temperature~0 and structured JSON output mode with enum validation.

\vspace{0.5em}
\small
\begin{verbatim}
You are a senior critical care physician extracting
clinical variables from a discharge summary for a
causal inference study on vasopressor therapy in sepsis.

You must return a JSON object with EXACTLY these 7 keys
and ONLY the allowed values listed below.

{
  "functional_status": one of ["independent",
    "partially_dependent", "fully_dependent", "unknown"],
  "mental_status": one of ["alert", "confused",
    "delirious", "obtunded", "comatose", "unknown"],
  "code_status": one of ["full_code", "DNR", "DNI",
    "comfort_measures_only", "unknown"],
  "infection_source": one of ["pulmonary", "abdominal",
    "urinary", "skin_soft_tissue", "bloodstream",
    "CNS", "other", "unknown"],
  "source_control": one of ["achieved", "not_achieved",
    "not_applicable", "pending", "unknown"],
  "family_support": one of ["actively_involved",
    "limited_involvement", "absent", "conflicted",
    "unknown"],
  "substance_use": one of ["none", "alcohol", "opioids",
    "stimulants", "multiple", "other", "unknown"]
}

DEFINITIONS:
- functional_status: BASELINE before acute illness.
- mental_status: At ICU ADMISSION, BEFORE sedation.
- code_status: Goals-of-care EARLY in admission.
- infection_source: Primary site causing sepsis.
- source_control: Procedurally controlled.
- family_support: Family/surrogate involvement.
- substance_use: History affecting hemodynamics.

RULES:
1. Extract ONLY pre-treatment / admission state.
2. If absent or ambiguous, use "unknown".
3. Return ONLY the flat JSON object.
\end{verbatim}
\normalsize
\paragraph{Open-source fine-tuning.}
To construct a gold-standard training set, we sample 3,200 discharge
summaries and independently extract the seven covariates using
GPT-4o, Gemini~2.5 Pro, and Claude Sonnet~4. For each covariate,
the final label is determined by majority vote across the three
models. We fine-tune Qwen3-14B-Instruct~\citep{Qwen3} on 2,560
consensus-labeled examples using LoRA (rank 16, learning rate
$2\times10^{-4}$, 3 epochs) and evaluate on the remaining 640.

\section{Implementation Details}
\label{app:implementation}

\paragraph{Propensity score estimation.}
All propensity scores are estimated via logistic regression. M2 uses
an L2 penalty with $C = 1.0$ to mitigate overfitting from the
50-dimensional TF-IDF input; all other methods use unregularized
logistic regression. LLM-extracted categorical covariates are
one-hot encoded before entering the propensity score model. All
structured tabular covariates enter without further transformation.

\paragraph{Matching.}
Matching uses 1:1 nearest-neighbor without replacement on the logit
propensity score with a caliper of 0.2 standard deviations.
Variance is estimated using Abadie--Imbens
heteroskedasticity-robust standard errors.

\paragraph{Inverse probability weighting.}
For M6, stabilized IPW weights use the marginal treatment
probability $\hat{P}(T{=}1)$ as the stabilization factor.
Propensity scores are clipped to $[0.01, 0.99]$ to enforce
positivity. Weights are truncated at the 1st and 99th percentiles
and renormalized within treatment groups. Variance is estimated
using robust sandwich standard errors.

\paragraph{BioClinicalBERT processing.}
We extract \texttt{[CLS]} token embeddings from
BioClinicalBERT~\citep{Alsentzer2019}. For notes exceeding 512
tokens, we apply a sliding window with mean pooling. The
768-dimensional embeddings are reduced to 50 dimensions via PCA.

\paragraph{Heterogeneous treatment effects.}
CausalForestDML~\citep{Wager2018} uses the five core LLM-extracted
covariates as candidate effect modifiers and the 26 tabular
covariates as nuisance confounders. Following Crump et
al.~\citep{Crump2009}, patients with extreme propensity scores
outside $[0.02, 0.98]$ are excluded, retaining 17,724 patients.

\paragraph{Doubly robust estimation.}
As a sensitivity analysis to assess robustness against
misspecification of the propensity score model, we estimate an
augmented inverse probability weighting (AIPW) treatment effect
using ForestDRLearner from the EconML library with 5-fold
cross-fitting. Both outcome and propensity score models are
implemented as gradient-boosted forests with 200 trees, and the
covariate set matches M4. Variance is estimated via honest forest
bootstrap.

\paragraph{E-value computation.}
E-values were computed on the risk ratio scale following VanderWeele
and Ding~\citep{VanderWeele2017}: given an estimated ATE on the risk
difference scale $\Delta$ and baseline risk $p_0 = 0.173$, we
convert to a risk ratio $\text{RR} = (p_0 + \Delta) / p_0$ and
then compute
$\text{E-value} = \text{RR} + \sqrt{\text{RR} \times (\text{RR} -
1)}$.

\section{Semi-Synthetic Experiments}
\label{app:synthetic}

Let $N = 21{,}859$ denote the total number of patients indexed by
$i$, $\sigma(z) = (1 + e^{-z})^{-1}$ denote the logistic function,
$\widetilde{\cdot}$ denote standardization to zero mean and unit
variance, and $\mathbb{1}[\cdot]$ denote the indicator function.
Continuous covariates are median-imputed before standardization;
categorical LLM covariates are one-hot encoded.

\paragraph{Treatment model.}
$T_i \sim \text{Bernoulli}\bigl(\sigma(s_i)\bigr)$ where $s_i$ is
the linear predictor:
\begin{align}
s_i = {} & \underbrace{\beta_1 \widetilde{\text{SOFA}}_i + \beta_2 \widetilde{\text{Age}}_i + \beta_3 \widetilde{\text{MAP}}_i + \beta_4 \widetilde{\text{Lactate}}_i}_{\text{tabular confounders}} \notag \\[4pt]
& + \underbrace{\gamma_1 \mathbb{1}[\text{full\_code}]_i + \gamma_2 \mathbb{1}[\text{CMO}]_i + \gamma_3 \mathbb{1}[\text{fully\_dep}]_i}_{\text{LLM confounders}} \notag \\[4pt]
& + \underbrace{\gamma_4 \mathbb{1}[\text{obtunded/coma}]_i + \gamma_5 \mathbb{1}[\text{pulmonary}]_i}_{\text{LLM confounders (cont.)}} + \beta_0
\end{align}
Coefficients $\beta$ correspond to tabular covariates and $\gamma$
to LLM covariates. The intercept $\beta_0$ is calibrated so that
the marginal treatment probability $\mathbb{E}[P(T=1)] \approx 0.10$
matches the observed treatment rate.

\begin{center}
\small
\begin{tabular}{lrl}
\toprule
\textbf{Parameter} & \textbf{Value} & \textbf{Role} \\
\midrule
$\beta_0$ (intercept) & $-1.50$ & calibrates $\mathbb{E}[\pi] \approx 0.10$ \\
$\beta_1$ (SOFA) & $+0.15$ & tabular \\
$\beta_2$ (Age) & $+0.08$ & tabular \\
$\beta_3$ (MAP) & $-0.20$ & tabular \\
$\beta_4$ (Lactate) & $+0.12$ & tabular \\
$\gamma_1$ (full code) & $+0.10$ & LLM \\
$\gamma_2$ (CMO) & $-0.50$ & LLM \\
$\gamma_3$ (fully dep.) & $+0.15$ & LLM \\
$\gamma_4$ (obtunded/coma) & $+0.20$ & LLM \\
$\gamma_5$ (pulmonary) & $+0.10$ & LLM \\
\bottomrule
\end{tabular}
\end{center}

\paragraph{Outcome model.}
$Y_i(t) \sim \text{Bernoulli}\bigl(\sigma(r_i + \tau \cdot t)\bigr)$,
where $r_i$ is the baseline risk linear predictor and $\tau$ is the
treatment effect on the logit scale:
\begin{align}
r_i = {} & \underbrace{\alpha_1 \widetilde{\text{SOFA}}_i + \alpha_2 \widetilde{\text{Age}}_i + \alpha_3 \widetilde{\text{MAP}}_i + \alpha_4 \widetilde{\text{Lactate}}_i}_{\text{tabular risk factors}} \notag \\[4pt]
& + \underbrace{\delta_1 \mathbb{1}[\text{CMO}]_i + \delta_2 \mathbb{1}[\text{fully\_dep}]_i + \delta_3 \mathbb{1}[\text{obtunded/coma}]_i}_{\text{LLM risk factors}} + \alpha_0
\end{align}
Coefficients $\alpha$ correspond to tabular covariates and $\delta$
to LLM covariates. The intercept $\alpha_0$ is calibrated so that
the marginal mortality rate $\approx 0.17$ matches the observed
rate.

\begin{center}
\small
\begin{tabular}{lrl}
\toprule
\textbf{Parameter} & \textbf{Value} & \textbf{Role} \\
\midrule
$\alpha_0$ (intercept) & $-1.20$ & calibrates mortality $\approx 0.17$ \\
$\alpha_1$ (SOFA) & $+0.20$ & tabular \\
$\alpha_2$ (Age) & $+0.10$ & tabular \\
$\alpha_3$ (MAP) & $-0.15$ & tabular \\
$\alpha_4$ (Lactate) & $+0.15$ & tabular \\
$\delta_1$ (CMO) & $+0.40$ & LLM \\
$\delta_2$ (fully dep.) & $+0.25$ & LLM \\
$\delta_3$ (obtunded/coma) & $+0.30$ & LLM \\
\bottomrule
\end{tabular}
\end{center}

We set $\tau = -0.05$ in the main experiment and $\tau = 0$ in the
null experiment.

\paragraph{Confounding structure.}
Both $s_i$ and $r_i$ depend on LLM covariates, so $\{Y(0),Y(1)\} \not\!\perp\!\!\!\perp T \mid \mathbf{X}^{\text{tab}}$. Methods omitting $\mathbf{X}^{\text{llm}}$ face unmeasured confounding by construction.

\paragraph{Protocol.}
We run $J = 200$ replications indexed by $j \in \{1, \ldots, 200\}$;
covariates are held fixed while $T_i$ and $Y_i$ are resampled in
each replication with seed $42 + j$. For each method $m$, we report
the following metrics, where $\hat\tau_m^{(j)}$ is the ATE estimate
from replication $j$ under method $m$ and
$\text{CI}_m^{(j)}$ is its 95\% confidence interval:
\begin{align}
\text{Bias}_m &= \bigl|\bar{\hat\tau}_m - \tau\bigr|, \quad \text{RMSE}_m = \sqrt{\tfrac{1}{J} \textstyle\sum_j (\hat\tau_m^{(j)} - \tau)^2} \\
\text{Coverage}_m &= \tfrac{1}{J} \textstyle\sum_j \mathbb{1}\bigl[\tau \in \text{CI}_m^{(j)}\bigr]
\end{align}
where $\bar{\hat\tau}_m = \tfrac{1}{J} \sum_j \hat\tau_m^{(j)}$ is
the mean estimate across replications.

\paragraph{Extraction noise.}
For each LLM covariate $m$ with $K_m$ categories, we simulate
misclassification by independently corrupting observations:
$\tilde{X}_{i,m} = X_{i,m}$ with probability $1 - p_{\text{flip}}$,
else $\tilde{X}_{i,m} \sim \text{Uniform}\{1, \ldots, K_m\}$.
Treatment and outcome are generated using true covariate values;
only the estimation step uses corrupted values. We evaluate
$p_{\text{flip}} \in \{0.05, 0.10, 0.20\}$.

\begin{figure}
\centering
\includegraphics[width=\columnwidth]{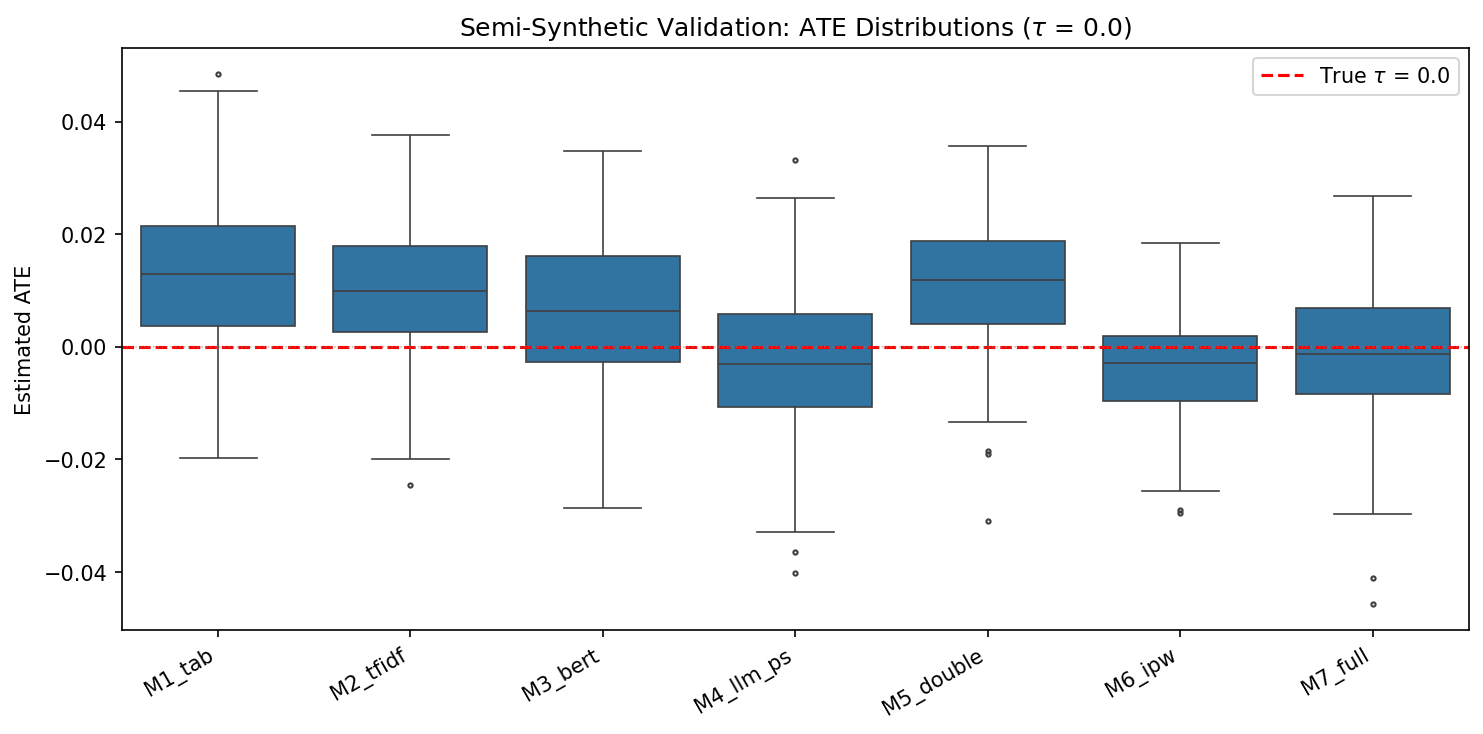}
\caption{ATE distribution under null effect ($\tau = 0$) across
200 simulations.}
\label{fig:null_boxplot}
\end{figure}

\paragraph{Results.}
Table~\ref{tab:semi_synthetic} reports per-method bias, RMSE, and
95\% CI coverage across 200 simulations under both a beneficial
treatment effect ($\tau = -0.05$) and a null effect ($\tau = 0$).
Table~\ref{tab:noise} summarizes robustness to simulated extraction
misclassification at error rates ranging from 0\% to 20\%, and
Figure~\ref{fig:null_boxplot} visualizes the distribution of ATE
estimates under the null scenario.

\vspace{0.5em}
\begin{center}
\begin{minipage}{\columnwidth}
\centering
\captionof{table}{Semi-synthetic validation over 200 simulations. Bias $= |\text{mean ATE} - \tau|$; Coverage $=$ fraction of 95\% CIs containing $\tau$.}
\label{tab:semi_synthetic}
\small
\begin{tabular}{lcccc}
\toprule
& \textbf{Mean ATE} & \textbf{Bias} & \textbf{RMSE} & \textbf{Coverage} \\
\midrule
\multicolumn{5}{l}{\textit{True $\tau = -0.05$}} \\
M1: Tabular PSM    & $-$0.0357 & 0.0143 & 0.0185 & 78.0\% \\
M2: TF-IDF PSM     & $-$0.0389 & 0.0111 & 0.0167 & 85.0\% \\
M3: BERT PSM       & $-$0.0418 & 0.0082 & 0.0149 & 91.5\% \\
M4: LLM PSM        & $-$0.0503 & 0.0003 & 0.0117 & 97.0\% \\
M5: Dual-caliper   & $-$0.0363 & 0.0137 & 0.0184 & 79.5\% \\
M6: LLM IPW        & $-$0.0512 & 0.0012 & 0.0080 & 95.5\% \\
M7: Full PSM       & $-$0.0495 & 0.0005 & 0.0118 & 96.0\% \\
\midrule
\multicolumn{5}{l}{\textit{True $\tau = 0$ (null effect)}} \\
M1: Tabular PSM    & $+$0.0123 & 0.0123 & 0.0175 & 81.5\% \\
M2: TF-IDF PSM     & $+$0.0102 & 0.0102 & 0.0158 & 87.0\% \\
M3: BERT PSM       & $+$0.0064 & 0.0064 & 0.0143 & 92.5\% \\
M4: LLM PSM        & $-$0.0027 & 0.0027 & 0.0126 & 95.5\% \\
M5: Dual-caliper   & $+$0.0110 & 0.0110 & 0.0158 & 90.0\% \\
M6: LLM IPW        & $-$0.0036 & 0.0036 & 0.0095 & 94.0\% \\
M7: Full PSM       & $-$0.0011 & 0.0011 & 0.0115 & 97.0\% \\
\bottomrule
\end{tabular}
\end{minipage}
\end{center}
\vspace{0.5em}

\vspace{0.5em}
\begin{center}
\begin{minipage}{\columnwidth}
\centering
\captionof{table}{Robustness to extraction noise. M4 bias remains below M1 bias at all error rates.}
\label{tab:noise}
\small
\begin{tabular}{cccc}
\toprule
\textbf{Error rate} & \textbf{M1 bias} & \textbf{M4 bias} & \textbf{M4 coverage} \\
\midrule
0\%  & 0.0143 & 0.0003 & 97.0\% \\
5\%  & 0.0143 & 0.0023 & 96.5\% \\
10\% & 0.0143 & 0.0045 & 93.0\% \\
20\% & 0.0143 & 0.0081 & 93.0\% \\
\bottomrule
\end{tabular}
\end{minipage}
\end{center}
\vspace{0.5em}

\section{Supplementary}
\label{app:figures}

\begin{table}[ht]
\centering
\small
\caption{Inter-model consensus rates on 3,200 sampled notes.
Consensus: two or more of three frontier models agree.}
\label{tab:agreement}
\begin{tabular}{lcc}
\toprule
\textbf{Covariate} & \textbf{Consensus} & \textbf{No consensus} \\
\midrule
Functional status   & 95.1\% &  4.9\% \\
Mental status       & 93.1\% &  6.9\% \\
Code status         & 97.0\% &  3.0\% \\
Infection source    & 96.5\% &  3.5\% \\
Source control      & 88.8\% & 11.2\% \\
Family support      & 99.4\% &  0.6\% \\
Substance use       & 98.8\% &  1.2\% \\
\midrule
Overall             & 95.5\% &  4.5\% \\
\bottomrule
\end{tabular}
\end{table}

\begin{figure}[ht]
\centering
\includegraphics[width=\columnwidth]{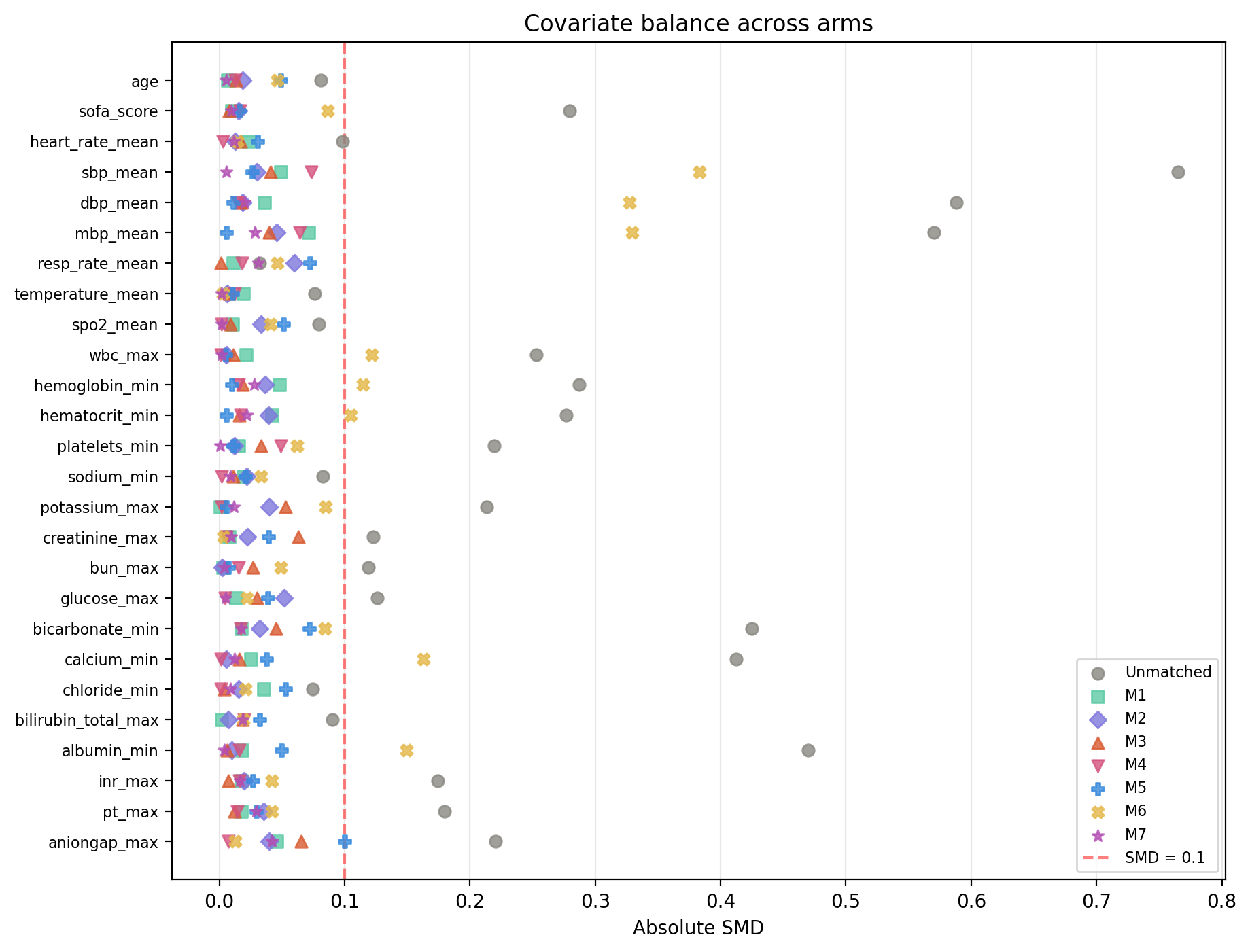}
\caption{Love plot of standardized mean differences across 26
tabular covariates. Gray circles show unmatched imbalances;
colored markers show post-adjustment balance for each method.
Dashed line indicates the SMD $= 0.1$ threshold.}
\label{fig:love}
\end{figure}

\begin{table}[ht]
\centering
\small
\caption{Post-matching SMD on LLM-extracted covariates.}
\label{tab:llm_balance}
\begin{tabular}{lccc}
\toprule
\textbf{Covariate} & \textbf{Unmatched} & \textbf{M1} & \textbf{M4} \\
\midrule
Functional: fully dependent & 0.066 & 0.040 & 0.007 \\
Functional: independent & 0.064 & 0.020 & 0.028 \\
Mental: obtunded & 0.114 & 0.087 & 0.018 \\
Mental: comatose & 0.088 & 0.095 & 0.012 \\
Code: full code & 0.233 & 0.227 & 0.002 \\
Code: CMO & 0.146 & 0.055 & 0.014 \\
Infection: pulmonary & 0.117 & 0.041 & 0.005 \\
Substance: alcohol & 0.067 & 0.038 & 0.004 \\
\midrule
\textbf{Mean (all)} & 0.087 & 0.058 & \textbf{0.013} \\
\textbf{Max (all)} & 0.288 & 0.227 & \textbf{0.046} \\
\bottomrule
\end{tabular}
\end{table}

\begin{figure}[ht]
\centering
\includegraphics[width=\columnwidth]{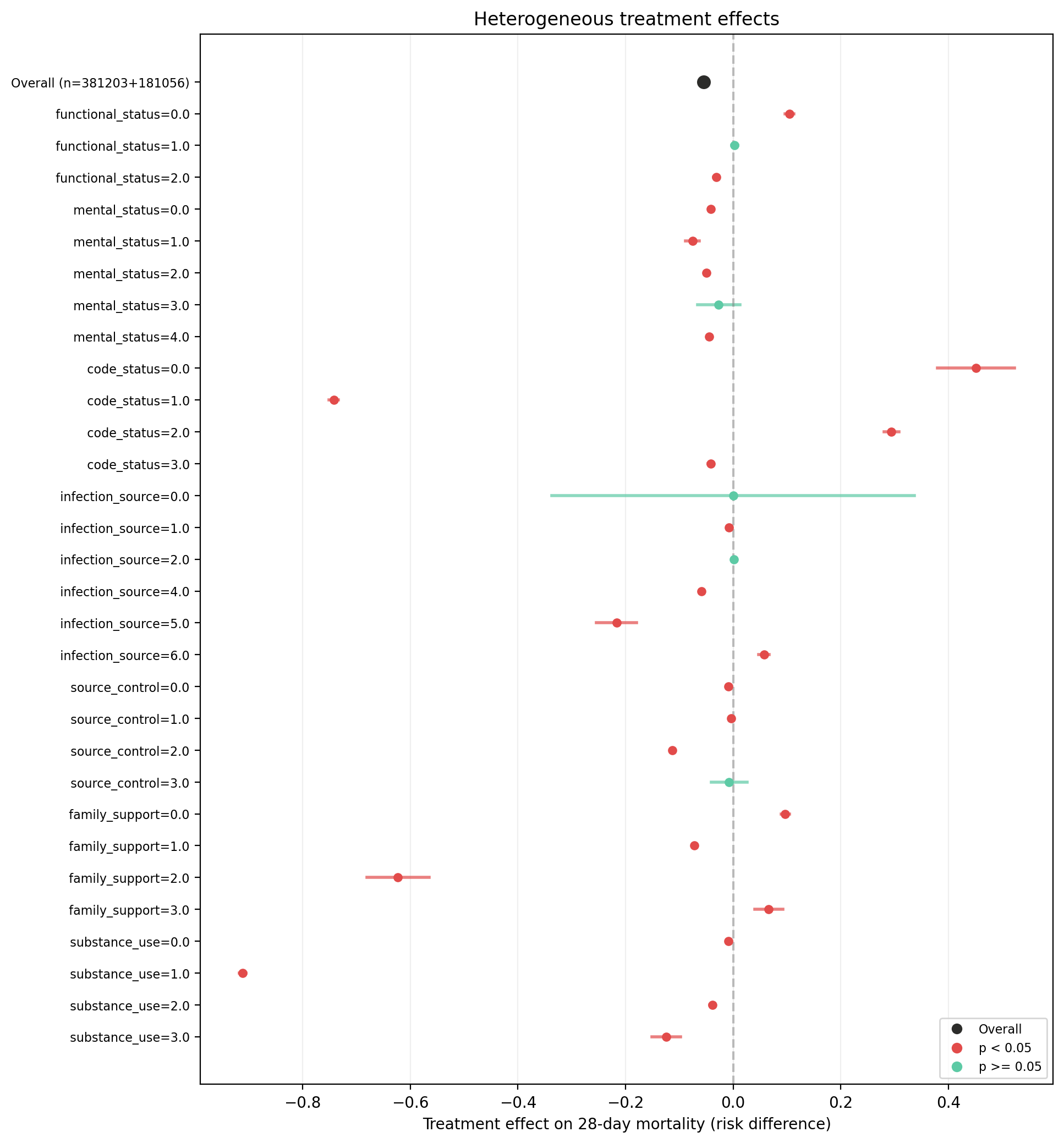}
\caption{Forest plot of ATE estimates across methods. Dashed line
indicates zero effect.}
\label{fig:forest}
\end{figure}

\begin{figure}[ht]
\centering
\includegraphics[width=\columnwidth]{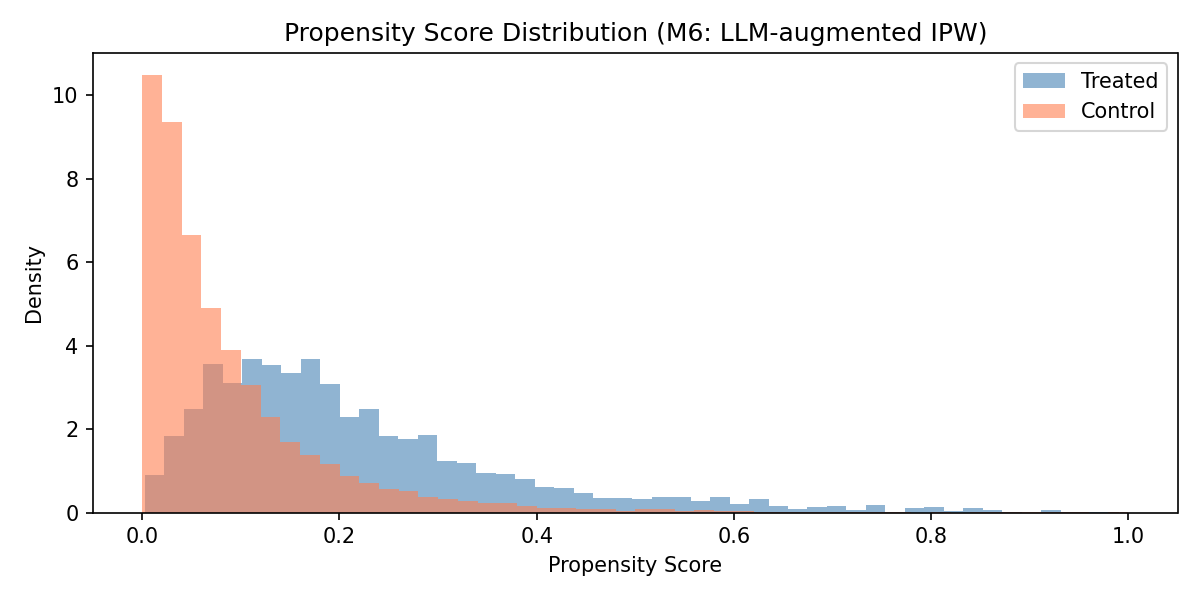}
\caption{Propensity score distributions for IPW (M6) by treatment
group.}
\label{fig:ps_dist}
\end{figure}

\begin{figure}[ht]
\centering
\includegraphics[width=0.85\columnwidth]{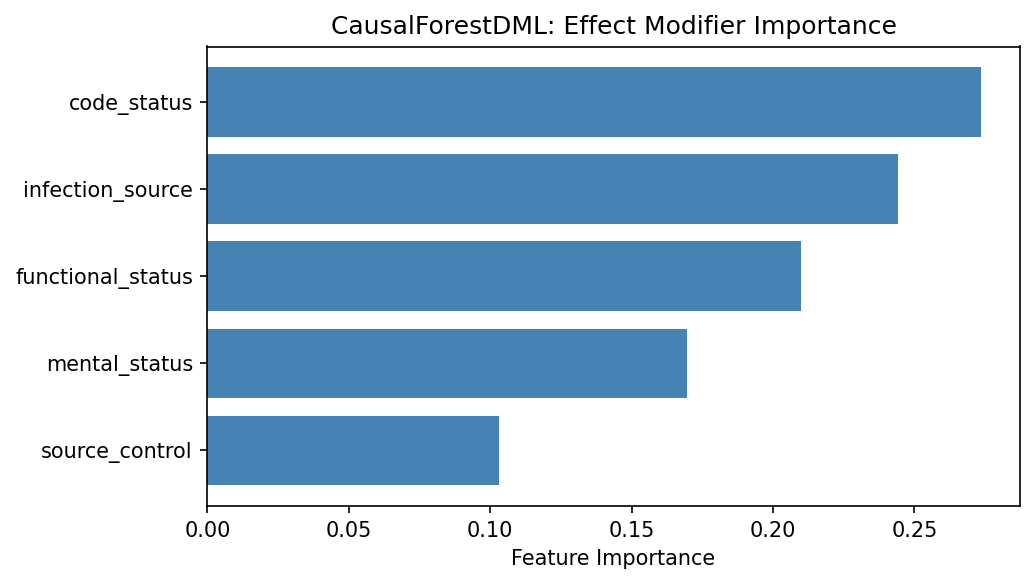}
\caption{Variable importance from CausalForestDML on the
PS-trimmed cohort of 17,724 patients. Code status, infection
source, and functional status collectively account for 74\% of
importance.}
\label{fig:varimp}
\end{figure}

\begin{table}[ht]
\centering
\small
\caption{Subgroup CATE by code status (PS-trimmed,
$n = 17{,}724$). All estimates are exploratory.}
\label{tab:hte_code}
\begin{tabular}{lccc}
\toprule
\textbf{Subgroup} & \textbf{CATE} & \textbf{95\% CI} & \textbf{$n$} \\
\midrule
Full code   & 0.011 & [$-$0.033, 0.054] & 6,005 \\
DNI         & 0.027 & [$-$0.026, 0.080] & 94 \\
CMO         & 0.025 & [$-$0.028, 0.079] & 1,293 \\
DNR         & 0.035 & [$-$0.058, 0.127] & 1,673 \\
Unknown     & 0.026 & [$-$0.046, 0.099] & 8,659 \\
\bottomrule
\end{tabular}
\end{table}

\begin{table}[ht]
\centering
\small
\caption{Subgroup CATE by functional status (PS-trimmed,
$n = 17{,}724$). All estimates are exploratory.}
\label{tab:hte_func}
\begin{tabular}{lccc}
\toprule
\textbf{Subgroup} & \textbf{CATE} & \textbf{95\% CI} & \textbf{$n$} \\
\midrule
Independent         & 0.005 & [$-$0.030, 0.040] & 3,769 \\
Partially dep.      & 0.021 & [$-$0.043, 0.084] & 6,430 \\
Fully dependent     & 0.032 & [$-$0.048, 0.112] & 2,021 \\
Unknown             & 0.031 & [$-$0.044, 0.106] & 5,504 \\
\bottomrule
\end{tabular}
\end{table}


\begin{figure}[ht]
\centering
\includegraphics[width=0.85\columnwidth]{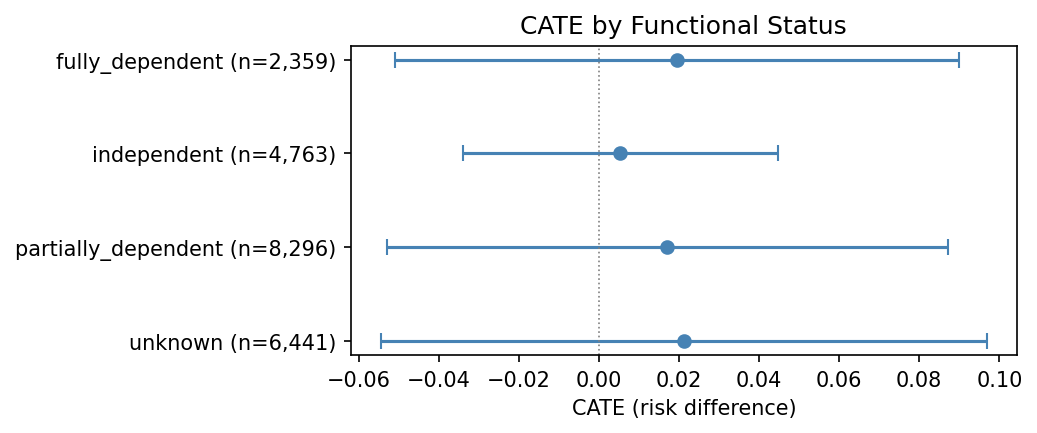}
\caption{Subgroup CATE estimates by functional status. All
confidence intervals include zero.}
\label{fig:hte_func}
\end{figure}

\begin{figure}[ht]
\centering
\includegraphics[width=0.85\columnwidth]{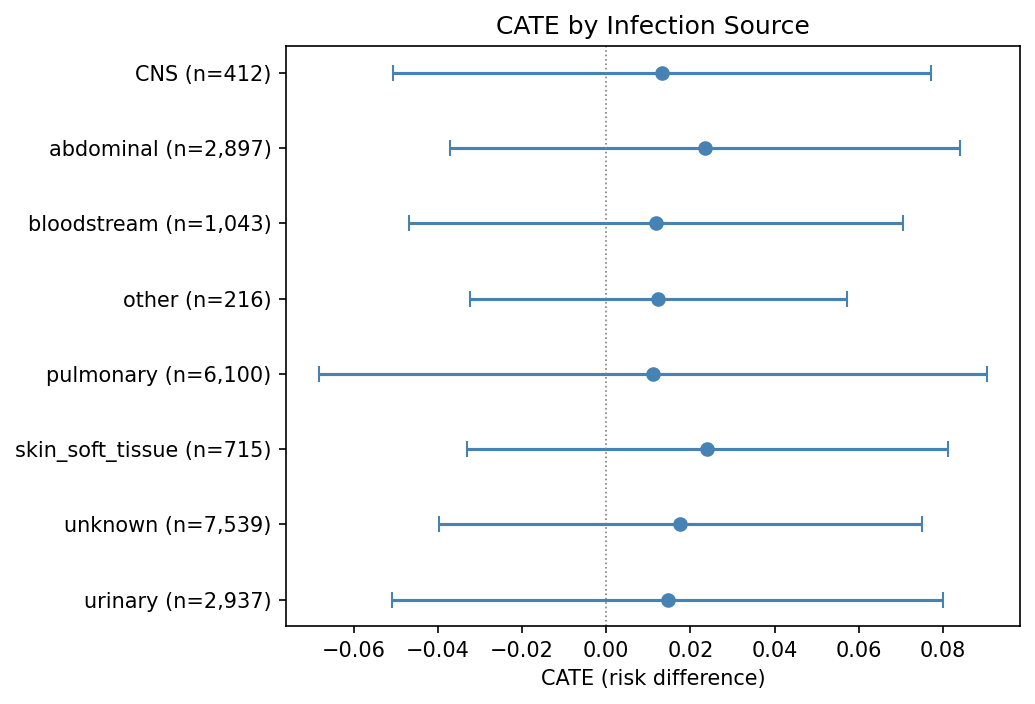}
\caption{Subgroup CATE estimates by infection source. All
confidence intervals include zero.}
\label{fig:hte_infection}
\end{figure}

\end{document}